%% file: templatePRIME.tex
\documentclass{article}

\usepackage{PRIMEarxiv}
\usepackage[numbers]{natbib}

\usepackage[table,xcdraw]{xcolor}
\usepackage{amsmath}
\usepackage{algorithm}
\usepackage{algpseudocode}
\usepackage{xcolor}
\usepackage{multirow}
\usepackage[utf8]{inputenc} 
\usepackage[T1]{fontenc}    
\usepackage[
    colorlinks=true,   
    pdfborder={0 0 1},  
    citebordercolor={0 1 0}, 
    linkbordercolor={0 1 0}, 
    urlbordercolor={0 1 0}   
]{hyperref}
\usepackage{url}            
\usepackage{booktabs}       
\usepackage{amsfonts}       
\usepackage{nicefrac}       
\usepackage{microtype}      
\usepackage{lipsum}
\usepackage{fancyhdr}       
\usepackage{graphicx}       
\graphicspath{{media/}}     

\newcommand{\bu}{\mathbf{u}}
\newcommand{\bn}{\mathbf{n}}

\newcommand{\E}{\mathcal{E}}

\newcommand{\grad}{\nabla}
\newcommand{\x}{\mathbf{x}}

\title{A Real-Time Event-Based Normal Flow Estimator}

\author{
  Dehao Yuan \\
  University of Maryland, College Park \\
  \texttt{dhyuan@umd.edu} \\
   \And
  Cornelia Fermüller \\
  University of Maryland, College Park \\
  \texttt{fermulcm@umd.edu} \\
}

\begin{document}
\maketitle

\vspace{-7pt}
\begin{abstract}
\input{src/s0_abstract}
\end{abstract}


\section{Introduction}
\label{sec:introduction}
\input{src/s10_introduction}

\section{Methodology}

\subsection{Problem Statement and Algorithm Overview}
\label{sec:overview}
\input{src/s21_overview}

\subsection{Optimized Local Events Encoding Using Random Fourier Features}
\label{sec:represent}
\input{src/s22_represent}

\subsection{Training}
\label{sec:training}

\input{src/s23_training}

\section{Runtime Analysis}
\label{sec:experiment}
\input{src/s30_runtime}

\section{Accuracy Analysis}
\label{sec:accuracy}
\input{src/s40_accuracy}

\section{Conclusion}
We introduce a high-accuracy, real-time normal flow estimator for event-based data. Our method employs a local event encoder to extract per-event features from spatiotemporal neighborhoods, followed by a lightweight two-layer perceptron for flow prediction. While the algorithm is based on \cite{yuan2024learning}, we reformulate it as a pooling operation, resulting in a significantly more efficient implementation. Moreover, the local event encoder is real-time and general-purpose, making it a valuable component for other event-based vision tasks.

\bibliographystyle{ieeetr}  
\bibliography{references}

\end{document}

%% file: src/s0_abstract.tex
This paper presents a real-time, asynchronous, event-based normal flow estimator. It follows the same algorithm as \textit{Learning Normal Flow Directly From Event Neighborhoods} \cite{yuan2024learning}, but with a more optimized implementation. The original method treats event slices as 3D point clouds, encodes each event’s local geometry into a fixed-length vector, and uses a multi-layer perceptron to predict normal flow. It constructs representations by multiplying an adjacency matrix with a feature matrix, resulting in quadratic time complexity with respect to the number of events. In contrast, we leverage the fact that event coordinates are integers and reformulate the representation step as a pooling operation. This achieves the same effect as the adjacency matrix but with much lower computational cost. As a result, our method supports real-time normal flow prediction on event cameras. Our estimator uses 1 GB of CUDA memory and runs at 4 million normal flows per second on an RTX 3070, or 6 million per second on an RTX A5000. We release the CUDA implementation along with a Python interface at \href{https://github.com/dhyuan99/VecKM_flow_cpp}{https://github.com/dhyuan99/VecKM\_flow\_cpp}.

%% file: src/s10_introduction.tex
Event-based motion field estimation is a fundamental problem for many downstream tasks, such as egomotion estimation \cite{ye2018unsupervised,zhu2019unsupervised} and motion segmentation \cite{barranco2014contour,mitrokhin2019ev,mitrokhin2020learning}. Moreover, an efficient and real-time algorithm for motion field estimation can significantly benefit robotics applications, including obstacle avoidance \cite{sanket2020evdodgenet} and visual servoing \cite{eguiluz2021fly}.

Existing works on event-based motion field estimation primarily focus on optical flow. Learning-based approaches, such as ERAFT \cite{gehrig2021raft}, rely on correlation volumes to match event patterns, while model-based methods like \cite{shiba2022secrets} use contrast maximization. Although performing well on static scenes, they both tend to degrade in accuracy when moving objects are present and are generally unable to achieve real-time performance \cite{liu2023tma, li2023blinkflow}. These limitations arise from the need to match large image neighborhoods to compute full flow vectors, which is computationally expensive. In dynamic scenes, occlusions near object boundaries further hinder accurate matching.

To avoid the limitations of optical flow, some methods use normal flow as a substitute. Normal flow is the component of optical flow parallel to the image gradient—it captures "half" of the full flow information. By sacrificing some directional detail, it can be estimated from a local image neighborhood without costly matching across large regions. Therefore, it can be computed both faster and more robust than optical flow. 

Normal flow has proven effective in downstream tasks such as egomotion estimation \cite{ren2024motion, lu2023event, li2024event}. However, most existing normal flow estimators are model-based, relying on plane fitting to the local space-time event surface \cite{benosman2013event, mueggler2015lifetime}. While these methods are efficient, they suffer from limited accuracy. To address this, \cite{yuan2024learning} proposed a learning-based normal flow estimator, achieving high accuracy but with high computational cost. This paper reformulates the algorithm in \cite{yuan2024learning}, which yields an efficient and accurate estimator.

We present a real-time, high-accuracy event-based normal flow estimator. It consists of a local event encoder that extracts per-event features from spatiotemporal neighborhoods, followed by a two-layer perceptron for flow prediction. While based on \cite{yuan2024learning}, our implementation is significantly more optimized and non-trivial, particularly in the design of the event-encoder. As a byproduct, we also provide a real-time, general-purpose local event encoder —it efficiently captures per-event features and can be reused in other event-based tasks. Our main contributions are:
\begin{itemize}
    \item We present a real-time event-based normal flow estimator with high accuracy.
    \item We develop a general-purpose, real-time local event encoder that can be applied to other event-based tasks.
\end{itemize}

%% file: src/s21_overview.tex
\textbf{Problem Statement.} The input to our normal flow estimator is a slice of events $\{(t_k, x_k, y_k)\}_{k=1}^n$, where $t_k\in[t_1, t_1+2\delta t]$, and $(x_k, y_k)$ are integer-valued pixel coordinates. The goal is to predict the \textit{generalized normal flow} for each event using its spatiotemporal neighborhood within the slice:
\begin{align}
    \text{CenteredLocalEvents}_k &=\Big\{\left[\begin{matrix}t-t_k\\x-x_k\\y-y_k\end{matrix}\right] : |x-x_k|\leq\delta x \text{ and } |y-y_k|\leq\delta y\Big\} \label{eqn:definition}\\
    \left[\hat{\bn}_x, \hat{\bn}_y\right]_k &= f(\text{CenteredLocalEvents}_k)\label{eqn:prediction}
\end{align}
Importantly, such formulation preserves the \textit{time-asynchronous} nature of events: even if two events occur at the same pixel location, their flow predictions can differ due to differences in their timestamps. 

\textbf{Generalized Normal Flow.} Normal flow $\bn$ is originally defined as the projection of optical flow $\bu$ onto the local image gradient direction:
\begin{equation}
    \bn=-\frac{\nabla I\cdot \bu}{||\nabla I||^2}\grad I
    \label{eqn:normal_flow_definition}
\end{equation}
Due to the formulation, normal flow and optical flow is connected through a normal flow constraint equation:
\begin{equation}
    \bn\cdot(\bu-\bn)=0
    \label{eqn:normal_flow_constraint}
\end{equation}
Note the normal flow $\bn$ is uniquely defined in Eqn. \eqref{eqn:normal_flow_definition}, while there are many $\bn$'s satisfying Eqn. \eqref{eqn:normal_flow_constraint}. In this paper, we refer to any flow vector that satisfies Eqn. \eqref{eqn:normal_flow_constraint} as a \textit{generalized normal flow}. Our goal is to predict the generalized normal flow defined by Eqn. \eqref{eqn:normal_flow_constraint}, rather than the specific form in Eqn. \eqref{eqn:normal_flow_definition}. This notion of generalized normal flow, as motivated in \cite{yuan2024learning}, allows supervision using ground-truth optical flow, and has been shown useful in downstream tasks such as egomotion estimation via a linear support vector classifier.

\textbf{Algorithm Overview.} To implement Eqn. \eqref{eqn:prediction}, we first convert the centered local events into a fixed-length vector representation, which is then passed through a two-layer neural network to predict the normal flow:
\begin{align}
    \texttt{emb}_k &= Encode(\text{CenteredLocalEvents}_k) \label{eqn:encode}\\
    \left[\hat{\bn}_x, \hat{\bn}_y\right]_k &= \texttt{MLP}(\texttt{emb}_k)
\end{align}
A naive approach to implement $Encode(\cdot)$ is using PointNet \cite{qi2017pointnet} and process each event independently. However, as elaborated at \cite{yuan2024linear}, this is highly inefficient due to the explicit construction of CenteredLocalEvents$_k$, and it does not reuse any computation when predicting normal flows for different events.

\cite{yuan2024learning} partially addresses this issue by using random Fourier features (RFF) to encode local events. Leveraging the translation equivariance of RFF \cite{yuan2024linear}, it enables feature reuse when estimating normal flow across different events (see Eqn. (4) in \cite{yuan2024learning}). However, the implementation relies on multiplying an adjacency matrix with a feature matrix, leading to quadratic time complexity with respect to the number of events.

In this paper, we exploit the fact that pixel coordinates are integers and reformulate the matrix multiplication as a pooling operation. This achieves the same effect as the adjacency matrix but with significantly lower computational cost. Algorithm \ref{alg:main} summarizes our optimized implementation, where \texttt{emb}$_k$ denotes the local event encoding of the $k$-th event. Although the implementation may appear ad hoc, Section \ref{sec:represent} provides a detailed explanation of the underlying mechanism.

\begin{algorithm}
\caption{Local Event Encoding Using Random Fourier Features.}
\begin{algorithmic}[1]  
\Require event slice $\{(t_k, x_k, y_k)\}_{k=1}^n$, camera resolution $(\texttt{W}, \texttt{H})$, dimension of event encoding \texttt{D}.
\Require time slice radius $\delta t$, pixel radius $\delta x$, $\delta y$ defined in Eqn. \eqref{eqn:definition}.
\Require compute event encodings at these event indices $K\subset[n]$ only.
\State \# An example of parameters: \texttt{W=640, H=480, D=64, $\delta t$=16ms, $\delta x$=$\delta y$=10\texttt{pxl}.}
\State \texttt{PixelEmbed = zeros(W, H, D)}
\State \texttt{PixelCount = zeros(W, H)}
\State \texttt{T = random\_normal(mean=0, var=25, size=D, seed=0)}
\State \texttt{X = random\_normal(mean=0, var=25, size=D, seed=1)}
\State \texttt{Y = random\_normal(mean=0, var=25, size=D, seed=2)}
\For{$k = 1$ to $n$}
    \State \texttt{PixelEmbed}$[x_k, y_k,:]$ \texttt{+=} $\displaystyle\exp\big(i\frac{t_k}{\delta t}\cdot \texttt{T}\big)$
    \State \texttt{PixelCount}$[x_k, y_k]$ \texttt{+=} 1
\EndFor
\For{$k\in K$}
\State $\texttt{emb}_k$ = \texttt{zeros(D)}
\State $\texttt{cnt}_k$ = 0
\For{$dx=-\delta x$ to $\delta x$}
\For{$dy=-\delta y$ to $\delta y$}
\State $\texttt{cnt}_k \texttt{ += PixelCount}[x_k+dx, y_k+dy]$
\State $\displaystyle\texttt{emb}_k \texttt{ += } \texttt{PixelEmbed}[x_k+dx, y_k+dy, :] * \underbrace{\exp\big(i\frac{dx}{\delta x}\cdot \texttt{X}\big) * \exp\big(i\frac{dy}{\delta y}\cdot \texttt{Y}\big)}_{\text{precomputed}}$
\EndFor
\EndFor
\State $\displaystyle\texttt{emb}_k \texttt{ = } \texttt{emb}_k * \exp(-i\frac{t_k}{\delta t}\cdot \texttt{T}) \div \texttt{cnt}_k$ \hfill \# this will be inputted to a two-layer network to predict normal flow.
\EndFor
\end{algorithmic}
\label{alg:main}
\end{algorithm}

%% file: src/s22_represent.tex
Algorithm \ref{alg:main} presents a seemingly ad hoc implementation of local event encoding. In this section, we provide the mathematical foundation behind this implementation. The explanation is distilled from \cite{yuan2024learning} and \cite{yuan2024linear}, and adapted to the context of this paper. The notations in this section are consistent with the variables in Algorithm \ref{alg:main}. To reiterate, our goal is to obtain an efficient encoding of centered local events:
\begin{equation}
    \texttt{emb}_k = Encode(\text{CenteredLocalEvents}_k)\label{eqn:encode2}
\end{equation}
This section is structured as following: Section \ref{sec:encode} introduces the formula that achieves Eqn. \eqref{eqn:encode2}. Section \ref{sec:pooling} reformulates this expression as a pooling operation, which matches the implementation in Algorithm \ref{alg:main}. Section \ref{sec:efficiency} explains its time complexity.

\subsubsection{$Encode(\text{CenteredLocalEvents}_k)$ using random Fourier features.}
\label{sec:encode}
Let $\E_0$ denote the centered local events around a single event $(t_0, x_0, y_0)$: 
\begin{equation}
    \E_0=\Big\{\left[\begin{matrix}t_k-t_0\\x_k-x_0\\y_k-y_0\end{matrix}\right] : |x_k-x_0|\leq\delta x \text{ and } |y_k-y_0|\leq\delta y\Big\}_{k=1}^n
\end{equation}
We can use Kernel Density Estimation \cite{pedregosa2025density} to model the density distribution of $\E_0$, where $\sigma^2$ is the bandwidth parameter of the KDE:
\begin{equation}
    g(t, x, y|\E_0)=\frac{1}{n}\sum_{k=1}^n \exp\Big(-\frac{1}{2\sigma^2}\Big[\big(\frac{t-t_k+t_0}{\delta t}\big)^2 + \big(\frac{x-x_k+x_0}{\delta x}\big)^2 + \big(\frac{y-y_k+y_0}{\delta y}\big)^2\Big] \Big) 
\end{equation}

$g(t,x,y|\E_0)$ is a descriptive expression of $\E_0$, but it is in a functional form which cannot be consumed by a neural network. If we can somehow convert $g(t,x,y|\E_0)$ into a vector representation, we are done with encoding the centered local events $\E_0$. Fortunately, we notice that $g(t,x,y|\E_0)$ can be represented by the following formula:
\begin{equation}
    \texttt{emb}_0=\frac{1}{n}\sum_{k=1}^n \exp\Big(i\left[\begin{matrix}(t_k-t_0)/\delta t\\(x_k-x_0)/\delta x\\(y_k-y_0)/\delta y\end{matrix}\right]^T\left[\begin{matrix}-&\texttt{T}&-\\-&\texttt{X}&-\\-&\texttt{Y}&-\end{matrix}\right]\Big)\in\mathbb{C}^\texttt{D}\label{eqn:emb}
\end{equation}
where $i$ is the imaginary unit. \texttt{T}, \texttt{X}, \texttt{Y} are three \texttt{D}-dimensional vectors following i.i.d. normal distribution with zero mean and variance $\sigma^2$ (see Line 4-6 in Algorithm \ref{alg:main}). 

To see \textbf{why Eqn. \eqref{eqn:emb} represents the kernel density estimation of $\E_0$}, an important observation is that $g(t,x,y|\E_0)$ can be reconstructed \textit{purely} from \texttt{emb}$_0$ by an inner product (proof in \cite{yuan2024linear} Lemma 1):
\begin{align}
\displaystyle
    \frac{1}{\texttt{D}}\Big\langle\texttt{emb}_0, \underbrace{\exp\Big(i\left[\begin{matrix}(t-t_0)/\delta t\\(x-x_0)/\delta x\\(y-y_0)/\delta y\end{matrix}\right]^T\left[\begin{matrix}-&\texttt{T}&-\\-&\texttt{X}&-\\-&\texttt{Y}&-\end{matrix}\right]\Big)}_{\text{does not depend on any $t_k, x_k, y_k$}}\Big\rangle &\rightarrow g(t,x,y|\E_0) \text{ as $D\rightarrow \infty$}
\end{align}
In other words, \texttt{emb}$_0$ is an equivalent representation of $g(t,x,y|\E_0)$, where $g(t,x,y|\E_0)$ models the density distribution of $\E_0$. Therefore, we conclude that \texttt{emb}$_0$ encodes $\E_0$, i.e. Eqn. \eqref{eqn:encode2} is achieved.

\subsubsection{Optimize with Pooling Operation}
\label{sec:pooling}
In the previous section, we discovered that Eqn. \eqref{eqn:emb} encodes the centered local events. We now rewrite it into pooling operation to compute it efficiently. First, we notice that Eqn. \eqref{eqn:emb} can be factorized into temporal and spatial components:
\begin{equation}
    \texttt{emb}_0=\frac{1}{n} \sum_{k=1}^n \underbrace{\exp(i\frac{t_k}{\delta t}\cdot \texttt{T}) * \exp(-i\frac{t_0}{\delta t}\cdot \texttt{T})}_{\text{temporal component}} * \underbrace{\exp\Big(i\left[\begin{matrix}(x_k-x_0)/\delta x\\(y_k-y_0)/\delta y\end{matrix}\right]^T\left[\begin{matrix}-&\texttt{X}&-\\-&\texttt{Y}&-\end{matrix}\right]\Big)}_{\text{spatial component}}
\end{equation}
First, $\displaystyle\exp(-i\frac{t_0}{\delta t}\cdot\texttt{T})$ is independent of $k$ and we move it out of the summation:
\begin{equation}
    \texttt{emb}_0=\exp(-i\frac{t_0}{\delta t}\cdot \texttt{T}) * \frac{1}{n} \sum_{k=1}^n 
    {\color{red}\exp(i\frac{t_k}{\delta t}\cdot \texttt{T})} * {\color{green}\exp\Big(i\left[\begin{matrix}(x_k-x_0)/\delta x\\(y_k-y_0)/\delta y\end{matrix}\right]^T\left[\begin{matrix}-&\texttt{X}&-\\-&\texttt{Y}&-\end{matrix}\right]\Big)}
    \label{eqn:13}
\end{equation}
Then since $(x_k, y_k)$ are integers, there may be multiple $t_k$'s corresponding to the same pixel coordinate $(x, y)$. In this case, the \textcolor{green}{green} part is identical for those $t_k$'s and only the \textcolor{red}{red} part is different. As a result, we can first compute 
\begin{align}
\texttt{PixelEmbed}[x,y]&=\sum_{\{k:(x_k,y_k)=(x,y)\}}{\color{red}\exp(i\frac{t_k}{\delta t}\cdot \texttt{T})} \\
\texttt{PixelCount}[x,y]&=\sum_{\{k:(x_k,y_k)=(x,y)\}}1 
\end{align}
and Eqn. \eqref{eqn:13} becomes 
\begin{align}
\begin{split}
    \texttt{emb}_0=\exp(-i\frac{t_0}{\delta t}\cdot \texttt{T}) \:* \frac{1}{n}\sum_{x=x_0-\delta x}^{x_0+\delta x}\:\sum_{y=y_0-\delta y}^{y_0+\delta y} \texttt{PixelEmbed}[x,y] * \underbrace{{\color{green}\exp\Big(i\left[\begin{matrix}(x-x_0)/\delta x\\(y-y_0)/\delta y\end{matrix}\right]^T\left[\begin{matrix}-&\texttt{X}&-\\-&\texttt{Y}&-\end{matrix}\right]\Big)}}_{\text{precomputed since $(x,y)$ are integers with finite ranges}}
    \label{eqn:16}
\end{split}
\end{align}
where 
\begin{equation}
    n=\sum_{x=x_0-\delta x}^{x_0+\delta x}\:\sum_{y=y_0-\delta y}^{y_0+\delta y} \texttt{PixelCount}[x,y]
    \label{eqn:17}
\end{equation}
Expanding Eqn. \eqref{eqn:16} \eqref{eqn:17} matches Algorithm \ref{alg:main}.This concludes the explanation of how Eqn. \eqref{eqn:emb} is reformulated as the pooling operation in Eqn. \eqref{eqn:16}.

\subsubsection{Efficiency}
\label{sec:efficiency}
To discuss the efficiency, we split the algorithm into two parts: 1. computing \texttt{PixelEmbed} and \texttt{PixelCount} (Line 7-10 in Algorithm \ref{alg:main}), 2. computing \texttt{emb}$_k$ with pooling operation (Line 11-21 in Algorithm \ref{alg:main}). Part 1 involves computing over all events, but the operation performed on each event is light. Part 2 involves heavier operations, but it only involves events whose normal flow is predicted. Consequently, the ultimate time complexity is 
\begin{equation}
    c\cdot num\_events + C\cdot num\_predicted\_flows
    \label{eqn:runtime}
\end{equation}
where $c\approx C(\delta x \delta y)^{-1}$. In Section \ref{sec:experiment}, we will provide the timing of each part and enrich the discussion.

%% file: src/s23_training.tex
The loss function and data sampling strategy are identical as \cite{yuan2024learning}. We train multiple recipes and release it for public use. Below summarizes the parameters and training set of each recipe:
\begin{table}[h]
\centering
\resizebox{0.75\textwidth}{!}{%
\begin{tabular}{@{}lcccccc@{}}
\toprule
\multicolumn{1}{c}{checkpoint name} &
  camera resolution &
  training set &
  $\delta x$ &
  $\delta y$ &
  $\delta t$ &
  \texttt{D} \\ \midrule
640x480\_32ms\_C64\_k8 &
  \multirow{4}{*}{(640, 480)} &
  \multirow{4}{*}{\begin{tabular}[c]{@{}c@{}}EVIMO2 \cite{burner2022evimo2} -- imo+sfm split,\\ recorded by Samsung mono\end{tabular}} &
  8 &
  8 &
  16ms &
  \multirow{4}{*}{64} \\
640x480\_32ms\_C64\_k10 &  &  & 10 & 10 & 16ms &  \\
640x480\_24ms\_C64\_k8  &  &  & 8  & 8  & 12ms &  \\
640x480\_24ms\_C64\_k10 &  &  & 10 & 10 & 12ms &  \\ \bottomrule
\end{tabular}%
}
\end{table}

%% file: src/s30_runtime.tex
We measure the execution time of two parts of the algorithm—Lines 7–10 and Lines 11–21—corresponding to the terms $c$ and $C$ in Eqn.~\eqref{eqn:runtime}. We evaluate the runtime on different GPU devices. Table \ref{tab:my-table} presents the results and Table \ref{tab:gpu} presents the GPU statistics. The tables allow users to estimate the approximate runtime for processing a given number of events. For instance, using an RTX 2080 Ti GPU with an input of 0.5 million events and computing 10,000 normal flows using a model with $\delta x = \delta y = 8$, the estimated runtime is:
\[\frac{0.5\text{M}}{115.63\text{M/s}} + \frac{10\text{k}}{4.25\text{M/s}} + \frac{10\text{k}}{27.55\text{M/s}}=7.04 \text{ms}\]

\begin{table}[h]
\centering
\caption{Runtime performance on different GPU devices, measured in processed events per second. M denotes million.}
\label{tab:my-table}
\resizebox{0.6\textwidth}{!}{%
\begin{tabular}{@{}lccccc@{}}
\toprule
                  & \multicolumn{5}{c}{$\delta x=\delta y=8$}                  \\ \cmidrule(l){2-6} 
                  & RTX 2080 Ti & RTX 3070 & RTX A4000 & RTX A5000 & RTX A6000 \\ \midrule
Line 7-10         & 115.63M     & 96.16M   & 114.87M   & 166.51M   & 166.74M   \\
Line 11-21        & 4.25M       & 3.09M    & 3.12M     & 5.61M     & 5.55M     \\
two-layer network & 26.67M      & 34.97M   & 36.23M    & 37.04M    & 31.45M    \\ \midrule
                  &             &          &           &           &           \\ \midrule
                  & \multicolumn{5}{c}{$\delta x=\delta y=10$}                 \\ \cmidrule(l){2-6} 
                  & RTX 2080 Ti & RTX 3070 & RTX A4000 & RTX A5000 & RTX A6000 \\ \midrule
Line 7-10         & 115.63M     & 96.16M   & 114.87M   & 166.51M   & 166.74M   \\
Line 11-21        & 2.70M       & 2.08M    & 2.09M     & 3.68M     & 3.68M     \\
two-layer network & 42.55M      & 33.22M   & 35.34M    & 27.78M    & 31.95M    \\ \bottomrule
\end{tabular}%
}
\end{table}

\begin{table}[h]
\centering
\caption{GPU statistics.}
\label{tab:gpu}
\resizebox{0.8\textwidth}{!}{%
\begin{tabular}{@{}lcccc@{}}
\toprule
                    & CUDA Core & GPU Memory  & Memory Bandwidth & FP32 Performance (TFLOPS) \\ \midrule
GeForce RTX 2080 Ti & 4352      & 11GB GDDR5X & 616GB/s          & 13.4                      \\
GeForce RTX 3070    & 5888      & 8GB GDDR6   & 448GB/s          & 20.3                      \\
RTX A4000           & 6144      & 16GB GDDR6  & 448GB/s          & 19.2                      \\
RTX A5000           & 8192      & 24GB GDDR6  & 768GB/s          & 27.8                      \\
RTX A6000           & 10752     & 48GB GDDR6  & 768GB/s          & 38.7                      \\ \bottomrule
\end{tabular}%
}
\end{table}

%% file: src/s40_accuracy.tex

Since the algorithm in this paper is fundamentally the same as that in \cite{yuan2024learning}, the flow prediction accuracy remains largely comparable. The main difference lies in the feature dimension: \cite{yuan2024learning} uses \texttt{D=384}, while we use a smaller \texttt{D=64} for improved efficiency. As shown in Table \ref{tab:evimo}, this reduction leads to a marginal drop in accuracy, while still outperforming other state-of-the-art optical flow estimators. Additionally, the higher accuracy reported in \cite{yuan2024learning} is partly due to its uncertainty quantification module, which filters out unreliable predictions and boosts overall accuracy. Besides quantitative results, the following links show flow predictions from each model evaluated on the EVIMO2-imo split:
\[ \href{https://drive.google.com/file/d/19AYypPS3cLnn9B0_bUY-0mEAdG6_4i7y/view?usp=drive_link}{\text{640x480\_32ms\_C64\_k8}}
\quad\href{https://drive.google.com/file/d/1bK1-69-9qlB0DNxI4ntkIH3aQ-QQUqtn/view?usp=sharing}{\text{640x480\_32ms\_C64\_k10}}
\quad\href{https://drive.google.com/file/d/1JHxR3t3o3D5Ts6oEHwpQZ2wj5yNXDIxF/view?usp=sharing}{\text{640x480\_24ms\_C64\_k8}}
\quad\href{https://drive.google.com/file/d/1AHMqFga7H36MRZcE4XejhSE-H-XzMZra/view?usp=sharing}{\text{640x480\_24ms\_C64\_k10}}
\]

\begin{table}[t]
\centering
\caption{Quantitative evaluation on EVIMO2 -- imo split \cite{burner2022evimo2}. M denotes MVSEC \cite{zhu2018multivehicle}. E denotes EVIMO2 \cite{burner2022evimo2}. D denotes DSEC \cite{gehrig2021raft}. Our new implementation is trained on EVIMO2.}
\label{tab:evimo}
\resizebox{\textwidth}{!}{%
\begin{tabular}{@{}lcccccccccccrrrrrr@{}}
\toprule
 &
   &
  \multicolumn{2}{c}{scene\_13\_00} &
  \multicolumn{2}{c}{scene\_13\_05} &
  \multicolumn{2}{c}{scene\_14\_03} &
  \multicolumn{2}{c}{scene\_14\_04} &
  \multicolumn{2}{l}{scene\_14\_05} &
  \multicolumn{2}{l}{scene\_15\_01} &
  \multicolumn{2}{l}{scene\_15\_02} &
  \multicolumn{2}{l}{scene\_15\_05} \\ \cmidrule(l){3-18} 
 &
  \multirow{-2}{*}{training set} &
  PEE &
  \% Pos &
  PEE &
  \% Pos &
  PEE &
  \% Pos &
  PEE &
  \% Pos &
  \multicolumn{1}{l}{PEE} &
  \multicolumn{1}{l}{\% Pos} &
  PEE &
  \%Pos &
  PEE &
  \%Pos &
  PEE &
  \%Pos \\ \midrule
 &
  M &
  \cellcolor[HTML]{EB948D}1.370 &
  \cellcolor[HTML]{FFFFFF}71.9\% &
  \cellcolor[HTML]{F2BAB5}2.406 &
  \cellcolor[HTML]{F9FDFB}90.6\% &
  \cellcolor[HTML]{EC9A93}1.356 &
  \cellcolor[HTML]{FFFFFF}69.5\% &
  \cellcolor[HTML]{ED9E97}1.458 &
  \cellcolor[HTML]{FFFFFF}64.6\% &
  \cellcolor[HTML]{ED9F99}2.186 &
  \cellcolor[HTML]{FFFFFF}67.1\% &
  \cellcolor[HTML]{F3BBB7}0.899 &
  \cellcolor[HTML]{FFFFFF}72.7\% &
  \cellcolor[HTML]{F2BAB6}0.980 &
  \cellcolor[HTML]{FFFFFF}67.1\% &
  \cellcolor[HTML]{F2B7B2}1.100 &
  \cellcolor[HTML]{FFFFFF}57.9\% \\
\multirow{-2}{*}{ERAFT \cite{gehrig2021raft}} &
  D &
  \cellcolor[HTML]{F7D3D0}0.843 &
  \cellcolor[HTML]{BDE5D2}88.9\% &
  \cellcolor[HTML]{FAE1DF}1.185 &
  \cellcolor[HTML]{9BD7BA}97.5\% &
  \cellcolor[HTML]{FFFBFB}0.517 &
  \cellcolor[HTML]{FFFFFF}88.1\% &
  \cellcolor[HTML]{FEF8F8}0.538 &
  \cellcolor[HTML]{FFFFFF}85.9\% &
  \cellcolor[HTML]{FCEBEA}0.908 &
  \cellcolor[HTML]{FFFFFF}86.3\% &
  \cellcolor[HTML]{FCEFEE}0.432 &
  \cellcolor[HTML]{E9F6F0}91.3\% &
  \cellcolor[HTML]{FBE5E4}0.541 &
  \cellcolor[HTML]{F0F9F5}90.9\% &
  \cellcolor[HTML]{F8D9D7}0.674 &
  \cellcolor[HTML]{FFFFFF}73.9\% \\ \midrule
 &
  M &
  \cellcolor[HTML]{F7D5D2}0.823 &
  \cellcolor[HTML]{F6FCF9}85.6\% &
  \cellcolor[HTML]{EDA09A}3.201 &
  \cellcolor[HTML]{C0E6D4}95.3\% &
  \cellcolor[HTML]{F2B6B1}1.111 &
  \cellcolor[HTML]{FFFFFF}86.3\% &
  \cellcolor[HTML]{EC9790}1.532 &
  \cellcolor[HTML]{FFFFFF}86.0\% &
  \cellcolor[HTML]{EA9088}2.445 &
  \cellcolor[HTML]{FFFFFF}82.2\% &
  \cellcolor[HTML]{F9DEDC}0.588 &
  \cellcolor[HTML]{FFFFFF}85.4\% &
  \cellcolor[HTML]{FAE4E2}0.556 &
  \cellcolor[HTML]{FFFFFF}87.7\% &
  \cellcolor[HTML]{F6CECB}0.811 &
  \cellcolor[HTML]{FFFFFF}68.0\% \\
\multirow{-2}{*}{TCM \cite{paredes2023taming}} &
  D &
  \cellcolor[HTML]{F9DBD8}0.774 &
  \cellcolor[HTML]{D9F0E5}87.3\% &
  \cellcolor[HTML]{F1B5B0}2.541 &
  \cellcolor[HTML]{C2E7D5}95.1\% &
  \cellcolor[HTML]{F7D2CF}0.872 &
  \cellcolor[HTML]{FFFFFF}87.8\% &
  \cellcolor[HTML]{F4C2BE}1.090 &
  \cellcolor[HTML]{FFFFFF}86.5\% &
  \cellcolor[HTML]{F3C0BB}1.640 &
  \cellcolor[HTML]{FFFFFF}84.1\% &
  \cellcolor[HTML]{FAE5E3}0.523 &
  \cellcolor[HTML]{FFFFFF}85.5\% &
  \cellcolor[HTML]{FBE7E5}0.528 &
  \cellcolor[HTML]{FFFFFF}87.9\% &
  \cellcolor[HTML]{F5C9C6}0.871 &
  \cellcolor[HTML]{FFFFFF}68.1\% \\ \midrule
MultiCM \cite{shiba2022secrets} &
  - &
  \cellcolor[HTML]{E8847B}1.509 &
  \cellcolor[HTML]{FFFFFF}53.2\% &
  \cellcolor[HTML]{E67C73}4.315 &
  \cellcolor[HTML]{FFFFFF}75.7\% &
  \cellcolor[HTML]{E67C73}1.611 &
  \cellcolor[HTML]{FFFFFF}79.2\% &
  \cellcolor[HTML]{E67C73}1.800 &
  \cellcolor[HTML]{FFFFFF}73.2\% &
  \cellcolor[HTML]{E67C73}2.768 &
  \cellcolor[HTML]{FFFFFF}72.9\% &
  \cellcolor[HTML]{F4C1BC}0.852 &
  \cellcolor[HTML]{FFFFFF}68.0\% &
  \cellcolor[HTML]{F6CCC8}0.802 &
  \cellcolor[HTML]{FFFFFF}66.2\% &
  \cellcolor[HTML]{F7D4D1}0.744 &
  \cellcolor[HTML]{FFFFFF}59.8\% \\ \midrule
PointNet \cite{qi2017pointnet} &
  E &
  \cellcolor[HTML]{F2BBB6}1.047 &
  \cellcolor[HTML]{CBEADC}88.1\% &
  \cellcolor[HTML]{FBE9E8}0.924 &
  \cellcolor[HTML]{96D5B7}97.7\% &
  \cellcolor[HTML]{F7D5D2}0.848 &
  \cellcolor[HTML]{7DCBA4}98.3\% &
  \cellcolor[HTML]{F8D6D3}0.892 &
  \cellcolor[HTML]{95D4B6}96.2\% &
  \cellcolor[HTML]{FAE3E1}1.053 &
  \cellcolor[HTML]{8DD1B0}96.6\% &
  \cellcolor[HTML]{F5CAC7}0.765 &
  \cellcolor[HTML]{8ED1B1}96.1\% &
  \cellcolor[HTML]{F7D1CD}0.752 &
  \cellcolor[HTML]{A7DCC3}95.1\% &
  \cellcolor[HTML]{F0B0AA}1.185 &
  \cellcolor[HTML]{E6F5EE}91.5\% \\ \midrule
PCA &
  - &
  \cellcolor[HTML]{E67C73}1.573 &
  \cellcolor[HTML]{CAEADA}88.2\% &
  \cellcolor[HTML]{F4C6C2}2.035 &
  \cellcolor[HTML]{FFFFFF}87.5\% &
  \cellcolor[HTML]{E78077}1.580 &
  \cellcolor[HTML]{FFFFFF}91.9\% &
  \cellcolor[HTML]{E77E75}1.784 &
  \cellcolor[HTML]{FAFDFC}90.3\% &
  \cellcolor[HTML]{F1B5B0}1.823 &
  \cellcolor[HTML]{FFFFFF}89.4\% &
  \cellcolor[HTML]{E67C73}1.467 &
  \cellcolor[HTML]{DCF1E7}92.1\% &
  \cellcolor[HTML]{E67C73}1.612 &
  \cellcolor[HTML]{FFFFFF}78.2\% &
  \cellcolor[HTML]{E67C73}1.821 &
  \cellcolor[HTML]{FFFFFF}84.7\% \\ \midrule
 &
  M &
  \cellcolor[HTML]{FAE2E0}0.713 &
  \cellcolor[HTML]{65C194}95.6\% &
  \cellcolor[HTML]{FFFEFE}0.269 &
  \cellcolor[HTML]{66C295}99.3\% &
  \cellcolor[HTML]{FBE9E7}0.676 &
  \cellcolor[HTML]{70C69C}98.8\% &
  \cellcolor[HTML]{FCEDEC}0.651 &
  \cellcolor[HTML]{72C69D}98.1\% &
  \cellcolor[HTML]{FDF1F0}0.806 &
  \cellcolor[HTML]{6EC59B}98.2\% &
  \cellcolor[HTML]{FCEBEA}0.470 &
  \cellcolor[HTML]{80CCA7}96.6\% &
  \cellcolor[HTML]{FDF0EF}0.433 &
  \cellcolor[HTML]{8BD1AF}95.8\% &
  \cellcolor[HTML]{FDF0EF}0.392 &
  \cellcolor[HTML]{B7E2CE}94.3\% \\
 &
  D &
  \cellcolor[HTML]{FDF1F0}0.590 &
  \cellcolor[HTML]{59BC8B}96.6\% &
  \cellcolor[HTML]{FFFFFF}0.230 &
  \cellcolor[HTML]{57BB8A}99.8\% &
  \cellcolor[HTML]{FDF4F4}0.575 &
  \cellcolor[HTML]{57BB8A}99.8\% &
  \cellcolor[HTML]{FDF0EF}0.625 &
  \cellcolor[HTML]{57BB8A}99.5\% &
  \cellcolor[HTML]{FFFFFF}0.567 &
  \cellcolor[HTML]{57BB8A}99.4\% &
  \cellcolor[HTML]{FDF4F3}0.391 &
  \cellcolor[HTML]{57BB8A}98.1\% &
  \cellcolor[HTML]{FFFDFD}0.298 &
  \cellcolor[HTML]{57BB8A}97.1\% &
  \cellcolor[HTML]{FCEDEC}0.424 &
  \cellcolor[HTML]{CAEADA}93.2\% \\
 &
  E &
  \cellcolor[HTML]{FFFCFB}0.497 &
  \cellcolor[HTML]{57BB8A}96.7\% &
  \cellcolor[HTML]{FEFAFA}0.399 &
  \cellcolor[HTML]{6AC397}99.2\% &
  \cellcolor[HTML]{FFFFFF}0.478 &
  \cellcolor[HTML]{66C295}99.2\% &
  \cellcolor[HTML]{FFFBFA}0.515 &
  \cellcolor[HTML]{65C194}98.8\% &
  \cellcolor[HTML]{FFFFFE}0.584 &
  \cellcolor[HTML]{67C295}98.6\% &
  \cellcolor[HTML]{FFFFFF}0.286 &
  \cellcolor[HTML]{57BB8A}98.1\% &
  \cellcolor[HTML]{FFFFFF}0.274 &
  \cellcolor[HTML]{63C093}96.8\% &
  \cellcolor[HTML]{FDF3F2}0.354 &
  \cellcolor[HTML]{6FC59B}95.5\% \\
\multirow{-4}{*}{VecKM\_flow \cite{yuan2024learning}} &
  M+D+E &
  \cellcolor[HTML]{FFFFFF}0.465 &
  \cellcolor[HTML]{5EBE8F}96.2\% &
  \cellcolor[HTML]{FFFDFD}0.308 &
  \cellcolor[HTML]{6AC397}99.2\% &
  \cellcolor[HTML]{FEF8F7}0.544 &
  \cellcolor[HTML]{64C193}99.3\% &
  \cellcolor[HTML]{FFFFFF}0.467 &
  \cellcolor[HTML]{65C194}98.8\% &
  \cellcolor[HTML]{FFFFFF}0.568 &
  \cellcolor[HTML]{69C297}98.5\% &
  \cellcolor[HTML]{FFFCFC}0.319 &
  \cellcolor[HTML]{60BF90}97.8\% &
  \cellcolor[HTML]{FFFDFD}0.300 &
  \cellcolor[HTML]{5BBD8D}97.0\% &
  \cellcolor[HTML]{FFFFFF}0.201 &
  \cellcolor[HTML]{57BB8A}95.7\% \\ \midrule
\multicolumn{2}{l}{640x480\_32ms\_C64\_k8} &
  \cellcolor[HTML]{FBE7E5}0.672 &
  \cellcolor[HTML]{A2DABF}90.8\% &
  \cellcolor[HTML]{FAE2E0}1.154 &
  \cellcolor[HTML]{C3E7D6}95.0\% &
  \cellcolor[HTML]{F4C6C2}0.977 &
  \cellcolor[HTML]{CFECDE}95.0\% &
  \cellcolor[HTML]{FAE0DE}0.784 &
  \cellcolor[HTML]{B8E2CE}94.3\% &
  \cellcolor[HTML]{F5C9C6}1.477 &
  \cellcolor[HTML]{B8E2CE}94.3\% &
  \cellcolor[HTML]{FBE9E7}0.493 &
  \cellcolor[HTML]{92D3B3}96.0\% &
  \cellcolor[HTML]{FCEFEE}0.441 &
  \cellcolor[HTML]{C5E8D7}93.5\% &
  \cellcolor[HTML]{FAE1DF}0.582 &
  \cellcolor[HTML]{D4EEE2}92.6\% \\
\multicolumn{2}{l}{640x480\_32ms\_C64\_k10} &
  \cellcolor[HTML]{FBE8E6}0.666 &
  \cellcolor[HTML]{BBE4D0}89.1\% &
  \cellcolor[HTML]{FAE5E3}1.055 &
  \cellcolor[HTML]{C6E8D8}94.8\% &
  \cellcolor[HTML]{F8D8D5}0.821 &
  \cellcolor[HTML]{CEECDD}95.0\% &
  \cellcolor[HTML]{FAE5E3}0.740 &
  \cellcolor[HTML]{B4E1CC}94.5\% &
  \cellcolor[HTML]{F6CCC9}1.425 &
  \cellcolor[HTML]{BDE5D2}94.0\% &
  \cellcolor[HTML]{FDF1F0}0.420 &
  \cellcolor[HTML]{9FD9BD}95.5\% &
  \cellcolor[HTML]{FDF0EF}0.428 &
  \cellcolor[HTML]{C0E6D4}93.8\% &
  \cellcolor[HTML]{FAE5E3}0.533 &
  \cellcolor[HTML]{CEECDE}92.9\% \\
\multicolumn{2}{l}{640x480\_24ms\_C64\_k8} &
  \cellcolor[HTML]{F9DDDA}0.760 &
  \cellcolor[HTML]{F4FBF8}85.7\% &
  \cellcolor[HTML]{FAE4E2}1.090 &
  \cellcolor[HTML]{D2EDE0}93.8\% &
  \cellcolor[HTML]{F5CAC6}0.945 &
  \cellcolor[HTML]{DDF2E7}94.4\% &
  \cellcolor[HTML]{F8D9D7}0.855 &
  \cellcolor[HTML]{C7E9D8}93.4\% &
  \cellcolor[HTML]{F5C7C3}1.511 &
  \cellcolor[HTML]{BEE5D3}93.9\% &
  \cellcolor[HTML]{FCEDEC}0.454 &
  \cellcolor[HTML]{AADDC5}95.0\% &
  \cellcolor[HTML]{FDF0EF}0.429 &
  \cellcolor[HTML]{D4EEE2}92.6\% &
  \cellcolor[HTML]{FAE2E0}0.564 &
  \cellcolor[HTML]{E0F3EA}91.9\% \\
\multicolumn{2}{l}{640x480\_24ms\_C64\_k10} &
  \cellcolor[HTML]{F9DEDC}0.746 &
  \cellcolor[HTML]{CCEBDC}88.0\% &
  \cellcolor[HTML]{FAE4E2}1.090 &
  \cellcolor[HTML]{C3E7D6}95.0\% &
  \cellcolor[HTML]{F4C3BF}1.002 &
  \cellcolor[HTML]{D2EDE0}94.8\% &
  \cellcolor[HTML]{F9DDDA}0.820 &
  \cellcolor[HTML]{C4E7D6}93.6\% &
  \cellcolor[HTML]{F5C8C4}1.498 &
  \cellcolor[HTML]{BFE5D3}93.8\% &
  \cellcolor[HTML]{FBE8E6}0.500 &
  \cellcolor[HTML]{9CD7BB}95.6\% &
  \cellcolor[HTML]{FCEDEB}0.468 &
  \cellcolor[HTML]{D5EEE2}92.5\% &
  \cellcolor[HTML]{F9DFDD}0.603 &
  \cellcolor[HTML]{D6EFE3}92.5\% \\ \bottomrule
\end{tabular}%
}
\end{table}